\pdfoutput=1

\documentclass[11pt,a4paper]{article}
\usepackage{acl2021}
\usepackage{times}
\usepackage{latexsym}

\usepackage{microtype}
\usepackage{graphicx}
\usepackage{comment}
\usepackage{breqn}
\usepackage{algorithm}
\usepackage{algpseudocode}
\usepackage{latexsym}

\title{Knowledge-Grounded Conversational Data Augmentation with Generative Conversational Networks}

\author{
  Yen-Ting Lin \\
  \texttt{ytl@ieee.org} \\
  \And
  Alexandros Papangelis \\
  \And
  Seokhwan Kim \\
  \texttt{\{papangea,seokhwk,hakkanit\}@amazon.com} \\
  \And
  Dilek Hakkani-Tur \\
}

\begin{document}
\maketitle
\begin{abstract}

While rich, open-domain textual data are generally available and may include interesting phenomena (humor, sarcasm, empathy, etc.) most are designed for language processing tasks, and are usually in a non-conversational format. In this work, we take a step towards automatically generating conversational data using Generative Conversational Networks, aiming to benefit from the breadth of available language and knowledge data, and train open domain social conversational agents. We evaluate our approach on conversations with and without knowledge on the Topical Chat dataset using automatic metrics and human evaluators. Our results show that for conversations without knowledge grounding, GCN can generalize from the seed data, producing novel conversations that are less relevant but more engaging and for knowledge-grounded conversations, it can produce more knowledge-focused, fluent, and engaging conversations. Specifically, we show that for open-domain conversations with 10\% of seed data, our approach performs close to the baseline that uses 100\% of the data, while for knowledge-grounded conversations, it achieves the same using only 1\% of the data, on human ratings of engagingness, fluency, and relevance. 
\end{abstract}

\section{Introduction}
Conversational Artificial Intelligence has progressed a lot in the recent past, partly due to advances in large pre-trained language models (PLM) and partly due to commercial conversational agents (Alexa, Siri, Cortana, Google Assistant, and others). It is evident, however, that many challenges still remain, such as handling idioms, humour, expressing empathy, processing unstructured knowledge, and so on. One big factor for this is the lack of large and rich conversational data that include these complex aspects of human communication. While the research community is making great efforts in collecting such data (e.g. empathetic dialogues \cite{rashkin2019towards}, persuasion \cite{wang-etal-2019-persuasion}, and others), these are still small compared to the amount of data needed to train deep neural networks. Furthermore, these expensive data collections usually target a single phenomenon at a time, and hence do not necessarily scale to the richness of human conversations. Another challenge for real world applications is privacy, preventing the use of much of the publicly available conversational data.

In this work, we take a first step into automatically generating conversational data from unstructured textual knowledge (e.g. web sources) using Generative Conversational Networks (GCN) \cite{papangelis2021generative}. GCN is a meta-learning method initially proposed for intent detection and slot tagging; we extend that approach and demonstrate that we can learn how to generate responses grounded in unstructured knowledge. Specifically, GCN learns how to generate labelled, diverse, and targeted data that are optimised with Reinforcement Learning (RL). This is achieved by using a generator model that produces new data which is used to train a separate learner model. The performance of the learner model is used as a reward signal to train the generator, so that over time the quality of the generated data increases. This reward signal can allow us to guide the data generation towards dimensions of interest, for example, knowledge-grounded, empathetic, or polite dialogues and can be derived from automatic metrics or human feedback if the system is deployed. In our case, the generator produces open-domain dialogues and the learner is a conversational agent that is trained on that data. Selecting an appropriate reward signal can be difficult, since we want to generate good quality dialogues that do not exist in the training data, but dialogue evaluation is a challenging open problem. We therefore investigate a combination of multiple metrics that capture different aspects: BLEU \cite{papineni2002bleu} and ROUGE \cite{lin2004rouge} to ensure some similarity with the reference data, BERTScore \cite{bert-score}\footnote{Data driven evaluation metrics tend to favor dialogues similar to the ones used during their training and we found that we cannot solely rely on such metrics.} to encourage good quality dialogues, and Knowledge F1\footnote{KF1 measures the token level F1 score between a knowledge piece and an utterance.} \cite{DBLP:conf/emnlp/0001PCKW21} to encourage knowledge integration. It should be noted that while the focus in this work is knowledge grounding in open-domain response generation, our approach is extensible to other conversational phenomena with appropriate reward signals.

Our main contributions are: a) we generate knowledge-grounded conversational data from unstructured textual knowledge (e.g.\ the kind of knowledge available on the web);  
b) we improve response generation quality over a baseline that uses fine-tuning on seed data, eliminating the need for additional human-human data collection; and c) we demonstrate improved performance on knowledge-grounded response generation on Topical Chat, as measured by KF1 and human evaluations.


\begin{figure*}[h]
    \centering
    \includegraphics[width=1\linewidth]{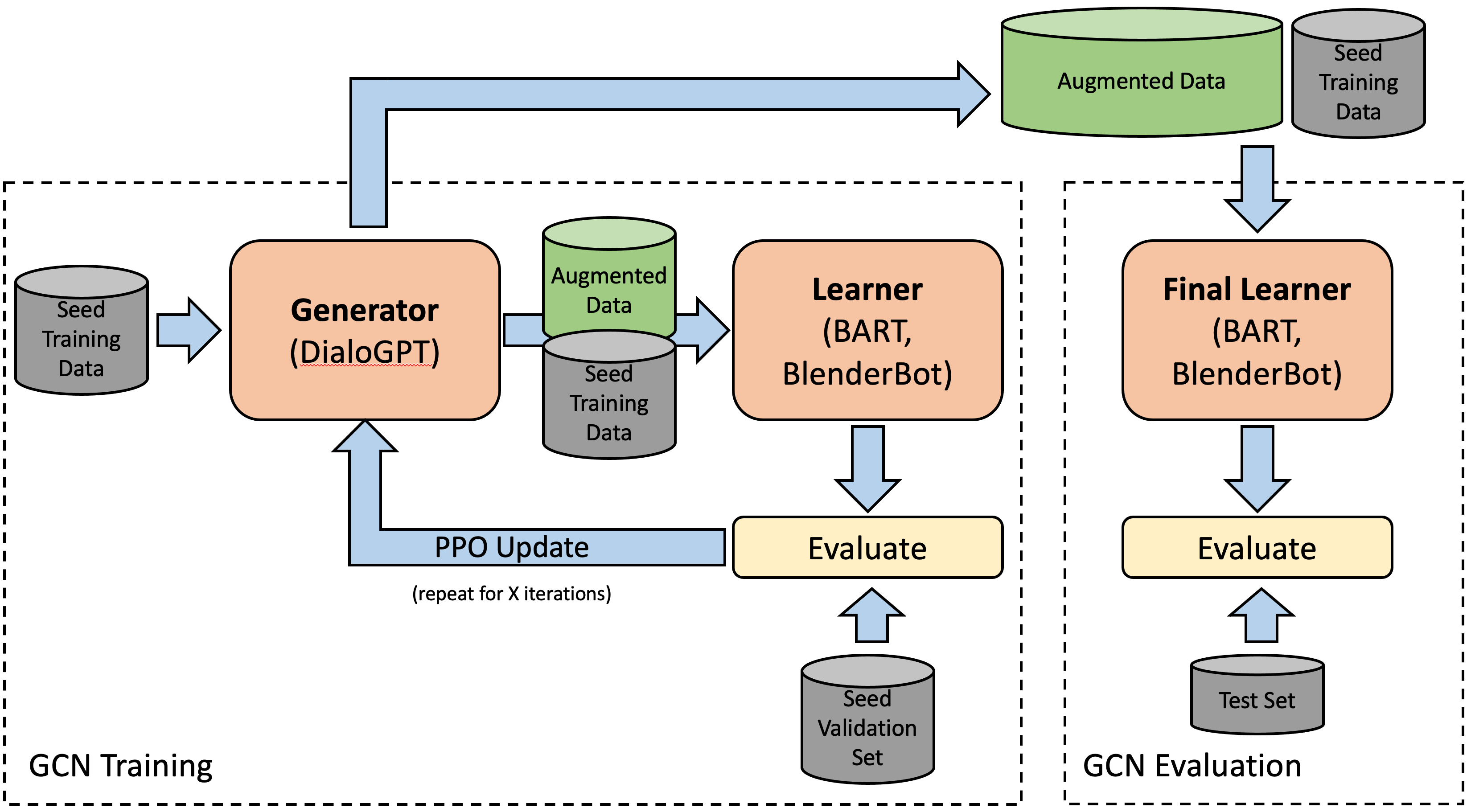}
    \caption{\small{The architecture of our approach using Generative Conversational Networks for knowledge-grounded dialogues. The generator is first fine-tuned with seed data and produces an augmented dataset and those data are used to train a learner. The performance of the learner on a held-out validation set (along with auxiliary metrics) is used as a reward to update the generator.}}
    \label{fig:gcn}
\end{figure*}

\section{Related Work}
\paragraph{Language Data Augmentation Approaches.}
There are a lot of recent works on data augmentation, but most of them are geared towards individual language processing tasks rather than training complete conversational agents. Due to lack of space we only mention the ones that are most relevant to our work. 

PROTODA \cite{kumar2021protoda} uses prototypical networks to augment data for intent classification while GenSF \cite{mehri2021gensf} uses DialoGPT \cite{zhang2020dialogpt} for zero-shot slot tagging; DINO \cite{DBLP:conf/emnlp/SchickS21a} uses PLM to generate data for semantic textual similarity; \citet{campagna2020zero} focus on zero-shot dialogue state tracking and use an abstract dialogue model to generate data. SOLOIST \cite{DBLP:journals/tacl/PengLLSLG21} uses a PLM fine-tuned on large dialogue corpora and is designed for transactional (goal-oriented) dialogues. \citet{mohapatra2020simulated} use PLM to train user simulators from crowd-generated conversations and their instructions. \citet{lin2021domain} train domain-independent user simulators for transactional dialogues. 
\citet{DBLP:conf/eacl/ChangSZDS21} augment data for Data-To-Text NLG by generating text in two steps: replacing values with alternatives and using GPT-2 to produce surface text. They then do automatic labelling and enforce cycle-consistency (make sure text can be generated from data and vice versa).
\citet{stahlberg2021synthetic} focus on data generation for Grammatical Error Correction and propose a method that can generate an erroneous sentence given a correct sentence and an error tag. \citet{DBLP:conf/emnlp/ChenY21} use data augmentation to improve out of scope (OOS) detection models. Specifically, they extract utterances from a different dataset than the one they are targeting that can be labelled as OOS and then do some smart filtering to select good candidates.
\citet{kim-etal-2021-neuralwoz} propose NeuralWOZ, a framework to generate dialogue state tracking data given goal descriptions and API calls. NeuralWOZ has a data generator and a data labeler that annotates the data. GCN does not need a separate labeler model and has the added option of being continually trained with RL. PromDA \cite{DBLP:conf/acl/0003XSHTGJ22} is a soft-prompt learning method for low-resource NLP tasks, that addresses the problem of overfitting (memorizing) when fine-tuning a PLM with a very small number of examples. The authors generate data for sequence classification and labelling. However, this approach is not tested on full dialogues which require significantly more context in the input.
\citet{bayer2022data} propose a three step method, where they first fine-tune a PLM and then generate new data-points by adjusting the temperature of the generation. They then filter the generated data by putting a threshold on embedding similarity with respect to the target class centroid. GCN uses RL to guide the generation process, alleviating the need for explicit post-processing. 
\citet{wang2022contrastive} present a data augmentation approach for aspect-based sentiment analysis that can generate data along two dimensions: aspects and polarity. The resulting data are then used in a contrastive learning setting to train a sentiment classifier. Similarly to other approaches, it is not clear how it would perform in knowledge-grounded dialogue generation, with large inputs (context and available knowledge).
For a more comprehensive review of data augmentation for language tasks, please see \cite{DBLP:conf/acl/FengGWCVMH21, DBLP:journals/corr/abs-2110-01852, DBLP:journals/coling/Sahin22}.

Regarding data augmentation for conversational agents, one of the most prominent methods is User Simulation \cite[e.g.]{schatzmann2007agenda,DBLP:conf/interspeech/AsriHS16,liu2018end,papangelis2019collaborative,DBLP:conf/sigdial/LinLHNGHFG21,shah2018building}. These approaches, however, have been designed to work with well-structured databases whereas we are concerned with grounding open-domain conversational responses in unstructured knowledge. $DG^2$ \cite{wu2021dg2} focuses on data augmentation for document-grounded dialogues, using Doc2Dial \cite{feng2020doc2dial}. The authors use an agent bot and a user bot to conduct simulated conversations and generate data. However, unlike GCN, the bots are not continually updated and may not generalise well to produce novel content. The code was not available for a direct comparison on our dataset, however, in the few-shot learning experiments, they demonstrate good performance with as little as 25\% of the data (869 Doc2Dial dialogues), whereas we demonstrate competitive performance by only using 1\% of the training data (86 Topical Chat dialogues).

\paragraph{Few-Shot Approaches.}
Another line of related work is based on few-/zero-shot transfer learning for dialogue tasks. Again due to space we only mention the most relevant works.
Earlier studies have focused on improving the generalizability of natural language understanding problems such as intent classification~\cite{chen2016zero} and slot filling~\cite{DBLP:conf/interspeech/BapnaTHH17,shah2019robust} for unseen labels or domains.
Then, focus was placed on other dialogue problems including dialogue state tracking~\cite{wu2019transferable,rastogi2020towards}, next action prediction~\cite{mosig2020star}, and natural language generation (NLG)~\cite{peng2020few}.
~\citet{DBLP:conf/interspeech/BapnaTHH17} and \citet{shah2019robust} utilized slot descriptions for improving the zero-shot slot filling performance.
~\citet{rastogi2020towards} used slot, intent, and task-specific API descriptions for schema-guided dialogue state tracking.
~\citet{mosig2020star} based on a structural schema in graph representations instead of textual descriptions for zero-shot action prediction and NLG.
~\citet{peng2020few} pre-trained on massive text data followed by dialog act labeled dialogue utterances.
~\citet{madotto2020language} used a large-scale pre-trained language model as a few-shot learner with task-specific prompting.
All the methods presented above, however, are geared towards specific tasks and are not shown to generalize to open-domain social or knowledge-grounded conversation.









\section{Notation}
We conduct experiments under two settings: conversations without explicit knowledge-grounding (we call them \emph{open-domain}) and knowledge-grounded conversations. \\
\subsection{Open-domain conversations}
We define a multi-turn conversation as a list of utterances: $U_1$, $U_2$, $...$, $U_N$ where $U_i$ is the utterance at turn $i$, and $N$ is the number of turns in the conversation.
Each utterance is composed of words $w_1$, $...$, $w_M$, where $M$ is the number of words in the utterance. 
Conversational agents are given a subset of the dialog context, for example the $t$ most recent turns $U_{N-t-1}$, $...$, $U_{N-1}$ and generate the response  $U_N$. \\
\subsection{Knowledge-grounded conversations}
To formulate knowledge-grounded responses, conversational systems need two steps (sometimes taken jointly): knowledge selection and response generation \cite{DBLP:conf/iclr/DinanRSFAW19}. 
The conversational agent should therefore first select relevant knowledge pieces from the sources provided with respect to the current dialog context and then generate a response that incorporates the selected knowledge. A knowledge piece in our case is defined as a fact consisting of one or more sentences (see Table \ref{tb:kg_conversations} for some examples). To select a knowledge retrieval method, we conducted preliminary experiments comparing TF-IDF, BM25, and BERTScore and we saw that the more sophisticated parsing and dense retrieval methods did not outperform TF-IDF.
We therefore represent conversation context and knowledge using TF-IDF vectors and utilize TF-IDF-based retrieval over documents as our knowledge selection mechanism. 
We select the most relevant knowledge using cosine similarity with the context $C=U_{N-t-1}, ..., U_{N-1}$:

\begin{equation}
    k_N = \operatorname*{argmax}_k\{\cos(t_C, t_k)\}
\end{equation}
where $t_C$ is the TF-IDF vector corresponding to the context and $t_k$ is the vector corresponding to knowledge piece $k$.
Knowledge-grounded conversational agents are given not only the dialog context $C$ but also the selected knowledge $k_N$ (or multiple pieces of knowledge as in our case) and are asked to generate a response $U_N$ that incorporates $k_N$.

\section{Generative Conversational Networks}
\label{sec:gcn}
GCN \cite{papangelis2021generative} (Figure \ref{fig:gcn}) consist of two models in a meta-learning architecture: a data generator and a learner. The generator creates a labeled dataset that is used to train a new learner (a conversational agent in our case) in a supervised fashion. The learner is then evaluated on an external validation set and its performance is used as a proxy for the quality of the dataset. This quality measure is used as a reward in a RL setup that trains the generator. Over time, the generator learns to create data of better and better quality, with respect to the learner's task, leading the learner to perform well. To avoid overfitting the validation set, we can limit the number of meta-iterations or include domain-independent performance metrics, such as fluency, perplexity, or even human feedback. When deployed, the generator is directly optimized on the test set (i.e. real interactions). Both models can be pre-trained with seed data, if available, and paired with reward estimation, GCN can be used for continuous learning from user feedback. This approach has been proven to work well for intent detection and slot tagging in goal-oriented conversations \cite{papangelis2021generative} and we here apply it to train social conversational agents. Different from Generative Adversarial Networks \cite{goodfellow2014generative}\footnote{A direct comparison with GAN approaches is out of scope for this work and we leave it for the future.} where the model tries to mimic the data, GCN models are guided by an external reward signal - that does not need to be differentiable - and can therefore generalize better. Depending on the optimization criteria, we can set the direction towards which the models will go, for example more polite conversations, more technical terminology, different dialect, knowledge grounding, and even directions that are not easily quantifiable (e.g. engagingness ratings from humans).

For open-domain conversations, as a proof of concept, we conduct few-shot experiments using 10\% of the data and for knowledge-grounded conversations which is the main focus of this work, we use 1\%, 5\%, and 10\% of the data; we call these the seed data ($D_{seed}$). At the beginning of training, we sample $D_{seed}$ from the data $D$, fine-tune the generator on $D_{seed}$ (see $G.train(D_{seed})$, line 4 in Algorithm \ref{alg:gcn}), and then start the outer loop meta-iterations. Along with the training data, we sample the corresponding percentage of validation data $D_{val}$. Once the training is complete, we spawn a new learner, train it on the seed and synthetic data, and evaluate it on $D_{test}$ which has been unseen so far. As described earlier, each meta-iteration has four phases: data generation, learner fine-tuning, learner evaluation, and generator update. Algorithm 1 summarizes the process.

\subsection{Data generation} 
In the first phase of the process, the generator $G$ is given some dialog context sampled from $D_{seed}$ and, in the knowledge-grounded condition, top-$m$ retrieved knowledge pieces $k$ from the TFIDF retriever. Specifically, we give the last two turns as context and the top-3 matching knowledge pieces, and ask the generator to predict the next system response. At each turn $i$, the context $C_i$ is used to retrieve relevant knowledge $k_i$ that is then used as input to the generator which produces the next turn response $U_i$:

\small
\begin{dmath}
    U_i = G(C_i, k_i) = \bigcup_{w=0}^n \{\textit{sample}(P_{LM}(w|w_{n-1},...,w_0,c_i,k_i) \}
\end{dmath}\normalsize
where $P_{LM}$ is the probability of the underlying language model generating each word $w$ of the response $U_i$, and \emph{sample} is the method we use to sample from the PLM, (greedy, nucleus, etc). This way, the generator produces a synthetic dataset $D_{synth}$ of size $L$, where each datapoint is a triplet of context $C_i$, knowledge $k_i$, and response $U_i$:

\begin{equation}
    D_{synth} = \{(C_i, k_i, U_i), i = 1, ..., L \}
\end{equation}
In essence, to create $D_{synth}$, instead of taking the human response from the data as a target, we use the generated response $U$ as a target and feed that along with $C$ and $k$ to fine-tune the learner.

\begin{algorithm}
\caption{GCN training procedure.}
\begin{algorithmic}[1]

\Procedure{Train}{$D_{seed}$, $D_{val}$, $D_{test}, \epsilon$}
    \State Initialize Generator $G$
    \If{$D_{seed}$}         
        \State $G$.train($D_{seed}$)
    \EndIf 
    \State Performance$_{meta} \leftarrow 0$
    \While{Performance$_{meta} < 1 - \epsilon$}
        \State $D_{synth} \leftarrow G$.generate()
        \State Sample and initialize new Learner $l$ 
        \State $l$.train($D_{seed} \cup D_{synth}$) 
        \State Performance$_{meta} \leftarrow l$.evaluate($D_{val}$)
        \State \Comment Performance$_{meta} \in [0, 1]$
        \State $G$.update(Performance$_{meta}$)
    \EndWhile  
    
    \State $D_{synth} \leftarrow G$.generate() 
    \State Sample and initialize new final Learner $L$
    \State $L$.train($D_{seed} \cup D_{synth}$)
    \State $L$.evaluate($D_{test}$)    \Comment{or other evaluator}
\EndProcedure
\end{algorithmic}
\label{alg:gcn}
\end{algorithm}

\subsection{Learner fine-tuning and evaluation} 
Since the learner's task is knowledge-grounded dialogue, it does not have access to the TFIDF retriever and, as $k$ may contain multiple relevant knowledge pieces, it will learn to perform its own implicit knowledge selection, not knowing what the exact knowledge piece used to produce $U$ was. 

At every iteration, we create a new learner (based on a pre-trained model) and fine-tune it on $D_{seed} \cup D_{synth}$ (see line 10 in Algorithm \ref{alg:gcn}). The knowledge-grounded learners are trained using a combination of cross entropy loss and knowledge retrieval score, specifically, Knowledge F1 (KF1) \cite{DBLP:conf/emnlp/0001PCKW21} which measures the F1 score between the produced utterance and the selected knowledge piece. The trained learner is then evaluated (see line 11 in Algorithm \ref{alg:gcn}) and a numerical reward is computed by combining several metrics.

\subsection{Generator update}
Following \cite{ziegler2019fine} and \cite{papangelis2021generative}, we use Proximal Policy Optimization (PPO) \cite{schulman2017proximal} with the following modified reward $R$ to train the generator using the learner's validation performance $r$:

\begin{equation}
    R(C, U) = r(C, U) - \beta \log\frac{G(U|C)}{G_{ref}(U|C)}
\end{equation}
where $C$ represents the context including the knowledge if applicable, $U$ represents the model's response, and $\beta$ is a constant that prevents $G$ from diverging too much from a reference generator $G_{ref}$.

In the open-domain condition, the generator uses multiple losses to calculate $r$: BLEU \cite{papineni2002bleu}, ROUGE-L \cite{lin2004rouge}, and BERTScore \cite{bert-score} which measure the similarity of the learner-produced utterance and the utterance in the data ($D_{seed}$ or $D_{synth}$). We evaluate each learner on the validation set $D_{val}$ and compute the above metrics using the human responses in $D_{val}$ as references. The weighted sum of the NLG metrics comprises the reward for the generator training. 
The weights were determined via grid search: $0.1$, $0.01$, $0.95$, for BLEU, ROUGE-L and BERTScore, respectively.
In the knowledge-grounded condition, we use a combination of BLEU-1 and KF1 (with weights $0.75$ for BLEU-1 and $0.25$ for KF1) as we found via grid search that it produced better results. 

After the meta-iterations are finished, we pick the best performing generator checkpoint (measured by the learners' performance on $D_{val}$ at each meta-iteration) and create a final synthetic set $D_{final\_synth}$ that is 5 times the size of the seed. We then create a new learner as our final learner (i.e. the conversational agent) and fine-tune it on $D_{seed} \cup D_{final\_synth}$ (lines 15-18 in Algorithm \ref{alg:gcn}). If $D_{final\_synth}$ is of good quality, we should expect the final learner to outperform the baseline, as it is trained with more data. The results presented next are all computed on the final learners, trained for 3 epochs, evaluated on $D_{test}$, and averaged over 3 runs (as are our baselines).

\begin{table*}[t]
\centering
\begin{tabular}{l||lll|cccc}
\hline
\textbf{Model} & \textbf{BLEU} & \textbf{Rouge(1/2/L)} & \textbf{BScore} & \textbf{Engaging.} & \textbf{Fluency} & \textbf{Relevance} & \textbf{Overall} \\
\hline
Data & - & - & - & 3.85 & 4.55 & 3.77 & 4.06 \\
BART (100\%) & 3.1 & 20.3/6.1/17.8 & 0.861 &  3.80 & 4.58 & 3.68 & 4.02 \\
\hline
BART (10\%) &\textbf{2.0} & \textbf{18.5/4.2/16.0} & {\bf 0.858} &  3.63 & 4.50  & {\bf 3.62} & 3.92 \\
GCN-RL & 1.1 & 15.0/2.1/12.6 & 0.850 &  3.70 & 4.47 & 3.47 & 3.88 \\
GCN+RL & 1.3 & 15.8/2.7/13.6 & 0.851 &  \textbf{3.79} & 4.49 & 3.58 & \textbf{3.96} \\
\hline
\end{tabular}
\caption{\small{Automatic and human evaluation results. Human evaluators rate responses on a scale of 1 to 5. BScore stands for BERTScore. Bold indicates statistically significant difference (t-test assuming unequal variance). BART (100\%) and BART (10\%) are BART trained on 100\% and 10\% of the data, GCN-RL is GCN without RL, and GCN+RL is GCN with RL training.}}
\label{tb:human_eval}
\end{table*}

\begin{table*}[!ht]
\centering
\begin{tabular}{l||lll|lll|lll}
\hline
&\multicolumn{3}{c}{{\bf 1\% data}} & \multicolumn{3}{c}{{\bf 5\% data}} & \multicolumn{3}{c}{{\bf 10\% data}} \\
\textbf{Model} & PPL & KF1 & BL-4 & PPL & KF1 & BL-4 & PPL & KF1 & BL-4 \\
\hline
BBs & 23.39 & 0.10 & 0.07 & 23.52 & 0.17 & 0.09 & 21.69 & 0.17 & 0.09 \\
GCN-RL & 26.47 & 0.15 & 0.08 & 24.54 & 0.18 & 0.09 & 23.11 & 0.18 & 0.09 \\
GCN+RL & 27.11 & {\bf 0.20} & 0.08 & 24.60 & {\bf 0.25} & 0.14 & 23.67 & {\bf 0.28} & 0.10 \\
\hline
\end{tabular}
\caption{\small{Results of automated evaluation on knowledge-grounded conversations. All models try to maximize KF1, and the baseline is the same model as the GCN learners (BBs: BlenderBot-small, 90M parameters).}}
\label{tb:kg_auto_eval}
\end{table*}

\begin{table}[t]
\centering
\resizebox{\linewidth}{!}{
\begin{tabular}{l|cccc}
\hline
\textbf{Model} & \textbf{Eng.} & \textbf{Flu.} & \textbf{Rel.} & \textbf{Avg} \\
\hline
Data & 3.74 & 3.98 & 3.57 & 3.76 \\ 
BBs (100\%) & 3.69 & 3.99 & 3.57 & 3.75 \\
\hline
BBs (1\%) & 3.64 & 3.86 & 3.42 & 3.64 \\
G-RL generator & 3.47 & 3.35 & 3.23 & 3.35 \\
G-RL learner & 3.58 & 3.85 & {\bf 3.48} & 3.64 \\
G+RL generator &  3.37 & 3.27 & 3.40 & 3.35 \\
G+RL learner& {\bf 3.73} & {\bf 3.97} & {\bf 3.48} & {\bf 3.73} \\
\hline
\hline
& \multicolumn{4}{c}{{\bf Wins Percentage}} \\
\hline
Combinations & \textbf{Base} & \textbf{G-RL} & \textbf{G+RL} & \textbf{Tie} \\
\hline
BBs VS G-RL & 40.0 & {\bf 44.3} & - & 15.7 \\
BBs VS G+RL & 44.7 & - & {\bf 47.7} & 7.6 \\
All 3 models & 29.3 & 25.7 & {\bf 45.0} & - \\
\hline
\end{tabular}}
\caption{\small{Human evaluation results (top) for knowledge-grounded conversations. Human evaluators rate responses with the same conversation context on a scale of 1 to 5. In a different evaluation (bottom), they were asked to choose the best response from two options. BBs: BlenderBot-small (90M), G-RL: GCN without RL, G+RL: GCN with RL.}}
\label{tb:kg_human_eval}
\end{table}

\section{Experiments}
To evaluate GCN as a data augmentation method for conversations with and without knowledge, we conduct few-shot experiments on Topical Chat (TC) \cite{gopalakrishnan2019topical}. TC is a set of human-human conversations, without explicitly defined roles for each participant, collected over Amazon Mechanical Turk. Each participant had access to a set of facts or articles with some conversations being symmetric (participants had access to the same knowledge) and some being asymmetric. All experiments were conducted on 2 Tesla V100 GPUs with 32GB memory each.

\subsection{Model ablations}
To quantify the effect of data augmentation and RL in both conditions, we train BART \cite{DBLP:conf/acl/LewisLGGMLSZ20} or BlenderBot-small (BBs)\footnote{90M parameters} \cite{DBLP:conf/eacl/RollerDGJWLXOSB21} models for no-knowledge and knowledge-grounded conversations respectively, under the following conditions:
\begin{itemize}
    \item {\bf Baseline (BART/BBs)}: In this condition, we train BART or BBs on the seed data. This will give us a lower bound on performance (if the augmented data is good, it should help performance).
    \item {\bf Data augmentation without RL (GCN-RL)}: In this condition, we pre-train a DialoGPT-small\footnote{117M parameters} \cite{zhang2020dialogpt} generator with the seed data, and use that to generate 5x more data. We then use the seed and generated data to train a final BART or BBs (learner) model depending on the task.
    \item {\bf Data augmentation with RL (GCN+RL)}: In this condition, we take the GCN-RL generator and iteratively update it using RL, as described in section 4. This is the full GCN framework. At the end of the meta-iterations, we take the best-performing generator and use it to create 5x more data. We use the seed and generated data to train a final BART or BBs model.
    \item {\bf Generator direct evaluation (G$\pm$RL generator)}: For the knowledge-grounded condition, in addition to the above three models, we evaluate the generator by having it directly interact with humans instead of generating data to train a learner. 
\end{itemize}

\subsection{Open-domain conversations}
For the open-domain conversations, we sample 10\% of TC as seed for GCN and use DialoGPT-small and BART as initial models for the generator and the learner, respectively. We compare the performance of the GCN learner and 3 baselines using automated metrics, and also conduct human evaluations. Our baselines are: BART trained with the same seed data (BART 10\%), BART trained with the entire training set (BART 100\%), and a GCN learner trained on seed and synthetic data but without updating the generator via RL (GCN-RL). Last, we also compare against the human responses that appear in the data (``Data" in Tables \ref{tb:human_eval} and \ref{tb:kg_human_eval}). 

\subsection{Knowledge-grounded conversations} 
For knowledge grounded conversations, we sample 1\%, 5\%, and 10\% of TC as seed data for GCN. 
Again we use DialoGPT-small as a generator but we use BBs as our learner. We compare the performance of GCN against similar baselines to the open-domain condition: BBs trained on the seed or the entire data, GCN without RL, human responses from the data, and we also evaluate the generators themselves if we were to use them directly as conversational agents (G$\pm$RL generator). Even though KF1 is the metric of choice in related work on knowledge-grounded conversations, we did not find works that report KF1 for TC.

\section{Results}
\subsection{Automatic evaluation} 
We report perplexity (PPL), BLEU-4 \cite{papineni2002bleu} with the ``method 7" smoothing function from \cite{chen2014systematic} as it has higher correlation with human ratings, and KF1. We calculate these metrics on the TC ``frequent" test set, (Tables \ref{tb:human_eval} and \ref{tb:kg_auto_eval}). In the open-domain condition, we see that BART 10\% outperforms GCN agents on all automated metrics. In knowledge-grounded conversations, we see that GCN+RL is able to incorporate more knowledge as evidenced by the higher KF1.

\subsection{Human evaluation}
Due to the intrinsic one-to-many property of conversation, reference-based metrics may not correlate with human ratings; our generated conversation may be appropriate for the dialogue context but different from the reference responses. For this reason, we also conduct human evaluation (following sub-section).
Human evaluators rate the output of the GCN learner, the baselines, and the ground truth. Specifically, they rate how engaging, fluent, and relevant each response is, on a scale from 1 to 5. We generate 1,000 samples for each condition using the same context and make sure we have 3 ratings per sample per condition. Tables \ref{tb:human_eval} (right) and \ref{tb:kg_human_eval} show the results of the evaluation, where we see that in the open-domain condition, the GCN learner produces engaging but less relevant conversations. This is likely because the model inserts facts or other output that is not entirely relevant, but is perceived as more engaging (e.g. information on a somewhat relevant subject, fun fact, etc.). Consistent with prior work, \cite{papangelis2021generative}, this shows that GCN can generalize from the data. When it comes to knowledge-grounded conversations, where GCN is explicitly trained to optimize KF1 (among other metrics), then relevance is indeed higher than the baseline. Overall, averaging the three metrics, GCN+RL outperforms BART 10\% and is close to BART 100\%'s performance. All models are outperformed by the human responses, which may be due to the size of our models or the number of training iterations. 

\begin{table}[ht]
\centering
\begin{tabular}{c|ccc}
\hline
\textbf{Iterations} & \textbf{PPL} & \textbf{KF1} & \textbf{BL-4} \\
\hline
1 & 30.8 & 0.146 & 0.179 \\
2 & 31.1 & 0.147 & 0.182 \\
3 & 30.7 & 0.146 & 0.186 \\
5 & 30.8 & 0.163 & 0.190 \\
10 & 27.1 & 0.238 & 0.085 \\
\hline
\end{tabular}
\caption{Performance of GCN+RL for varying number of meta-iterations. Here, we generate 3x the seed data and use 1\% of TC.}
\label{tb:meta_iter}
\end{table}

\begin{table}[ht]
\centering
\begin{tabular}{c|ccc}
\hline
\textbf{Data Mult.} & \textbf{PPL} & \textbf{KF1} & \textbf{BL-4} \\
\hline
1 & 26.5 & 0.201 & 0.082 \\
2 & 27.4 & 0.213 & 0.084 \\
3 & 28.6 & 0.17 & 0.083 \\
5 & 22.2 & 0.25 & 0.154 \\
10 & 22.9 & 0.27 & 0.106 \\
\hline
\end{tabular}
\caption{Performance of GCN+RL for varying size of generated data (as a multiplier of the seed). Here, we do 5 meta-iterations and use 1\% of TC.}
\label{tb:data_mult}
\end{table}

\begin{table}[ht]
\centering
\begin{tabular}{c|ccc}
\hline
\textbf{Data \%} & \textbf{BBs} & \textbf{GCN-RL} & \textbf{GCN+RL} \\
\hline
1\% & 8.1\% & 17.4\% & 25.1\% \\
5\% & 8.5\% & 12.1\% & 24.5\% \\
10\% & 5.9\% & 9.2\% & 13.6\% \\
\hline
\end{tabular}
\caption{Out-Of-Vocabulary (OOV) rates for various seed percentages.}
\label{tb:oov}
\end{table}

For knowledge-grounded conversations (Table \ref{tb:kg_human_eval}) we see that GCN+RL produces more engaging and fluent conversations and overall outperforms both baselines while again being close to BBs trained on all the data. In pair-wise comparisons, GCN+RL is generally preferred more than the other models. Overall, for the GCN conditions, given that we generate 5x the seed data, the total amount of data is about 6\% of the size of TC and our results show that the generated data is indeed of high quality, since the same model (BlenderBot-small) using the generated data performs close to the one that uses 100\% of the human-human data and close to the data itself. It should be noted that GCN achieves this performance using small models (in the order of 100M parameters each). 

In Figure \ref{fig:amt_setup} in the appendix, we show the Amazon Mechanical Turk setup that we used during our human evaluations.

\subsection{Generated data diversity}
In this section we further analyze the performance of GCN, specifically its performance with respect to the number of meta-iterations (Table \ref{tb:meta_iter}) and the amount of generated data (Table \ref{tb:data_mult}). In Table \ref{tb:meta_iter}, we see that KF1 increases as we have more meta-iterations, meaning that the generator actually leads the learner to learn to produce more knowledgable responses. BLEU naturally drops as these more knowledgable responses may not appear in the data. We observe similar trends in Table \ref{tb:data_mult}, where we vary the amount of synthetic generated data (as a multiplier of the size of the seed data). Regarding data diversity, Table \ref{tb:oov} presents out of vocabulary rates for all three conditions when using 1\%, 5\%, and 10\% of the data as seed. Higher rates mean more diversity but may also mean that the generated data is farther from the seed data. Paired with the results in Tables 1-3, however, we can see that GCN+RL produces more diverse data that are still in-domain and useful.

\subsection{Examples}
In Table \ref{tb:no_kg_conversations}, we provide example responses for conversations without knowledge grounding, and in Table \ref{tb:kg_conversations} we show example responses for knowledge-grounded conversations. We see that in both cases GCN+RL tries to insert knowledge or fun facts and that appear to be received well by the human judges. However, since there are no knowledge-grounding abilities, the model hallucinates and this is what likely drives relevance ratings down. In the knowledge-grounded example, we see that GCN+RL can use the knowledge pieces more effectively than the other models.

\begin{table*}[ht]
\resizebox{\textwidth}{!}{
\begin{tabular}{c|l}

\multicolumn{2}{c}{{\bf Context}} \\
\hline
\hline
& Hello, do you watch football? \\
\cline{2-2}
& Hi, yes. I love football! I did not realize the season lasted 17 weeks! \\
\cline{2-2}
& Yeah pretty long. I guess with 32 teams playing it takes that much time \\
\cline{2-2}
& Did you see they only rest one weekend during the season? \\
\cline{2-2}
& That is crazy, so much work goes in there. Did you know U of Iowa painted their visitor locker \\
& room pink? \\
\cline{2-2}
& Oh yes,that is so funny. they really want to intimidate the other team, but is more like a joke \\
\cline{2-2}
& Wow, what an idea. Did you know bowlers used to make more than football players in the 60s \\
\multicolumn{2}{c}{{\bf Responses}} \\
\hline
\hline
Baseline & that's a good point. i wonder how many of the top bowlers make twice as much as top football stars. \\
& do you have a favorite\\
\hline
GCN-RL & that's a good point. i'm not a big fan of football, but i do like to watch it. \\
& do you like football?\\
\hline
GCN+RL & i didn't know that, that's interesting. did you know there are more than 30 million players in the nfl? \\
& i wonder how many of those players are still playing?\\

\end{tabular}}
\caption{Example responses for open-domain conversations. The models here are {\bf not} trained to optimize knowledgable responses and therefore may hallucinate. This may be the reason for lower relevance scores for GCN+RL (Table \ref{tb:human_eval}); however, GCN+RL's tendency to provide fun facts is what may be leading to more engaging responses.}
\label{tb:no_kg_conversations}
\end{table*}

\begin{table*}[ht]
\centering
\resizebox{\textwidth}{!}{
\begin{tabular}{c|l}

\multicolumn{2}{c}{{\bf Context}} \\
\hline
\hline
& Did you know that the University of Iowa's locker room is painted pink? I wonder why? \\
\cline{2-2}
& I think I did hear something about that.  I imagine it is an attempt to psych the other team out. \\
\cline{2-2}
& So, it would be in the visiting team's locker room but not their own? \\
\multicolumn{2}{c}{{\bf Retrieved Knowledge (subset)}} \\
\hline
\hline
& the University of Iowa's locker room for visiting football teams is completely painted pink
in the\\
& 1960's top bowlers made twice as much as top football stars \\
\cline{2-2}
& Football is a family of team sports that involve, to varying degrees, kicking a ball [...] \\
\cline{2-2}
& in 39 out of 50 states, the highest paid state employee is either a college football or basketball coach. \\
\multicolumn{2}{c}{{\bf Responses}} \\
\hline
\hline
Baseline & i think it would be in the visiting team's locker room. i don't know why they would want to psych \\
& the other team out\\
\hline
GCN-RL & that's interesting. i've never heard that before. do you have a favorite team?\\
\hline
GCN+RL & i'm not too familiar with iowa's locker room for visiting football teams, but i do know that they're \\
& the highest paid employees in the state of iowa. do you know why they are paid so much? \\

\end{tabular}}
\caption{Example responses for knowledge-grounded conversations. In this setting, models are trained to optimize knowledgable responses which explains in part why GCN+RL is preferred in human judgements (Table \ref{tb:kg_human_eval}).}
\label{tb:kg_conversations}
\end{table*}

\section{Conclusion}
We presented a meta-learning method based on GCN to generate conversational data grounded on unstructured textual knowledge such as what can be found on the web. We show that given a small seed (1\% of TC or 86 examples) our approach can generate high quality data that can be used to train a competitive conversational agent able to do knowledge selection and grounding. Lower reference-based metric scores (Table \ref{tb:human_eval}, left and Table \ref{tb:kg_auto_eval}) indicate that our method generates diverse data. Based on the good scores in the human evaluation (Table \ref{tb:human_eval}, right and Table \ref{tb:kg_human_eval}), we see that the generated data is also useful for the task at hand, a result in line with the results on intent detection and slot tagging in \cite{papangelis2021generative}. As future work, we are conducting thorough analyses and evaluations of the generated data, exploring more model architectures, and are working on more conversational phenomena.

\bibliographystyle{acl_natbib}
\bibliography{acl2021}

\clearpage

\appendix
\section{Limitations and Risks}
Given that GCN relies on Reinforcement Learning, it may be difficult to tune for more complex applications. Also, due to its meta-learning nature it can be computationally expensive, although we show it here working with small versions of each model.

As with any machine learning approach - especially ones that rely on PLMs - there are inherent risks as to the language that will be produced, which may be biased in many ways. Our approach works with any machine learning model and therefore inherits the risks of the models used. Reinforcement Learning can also play a role in learning biased models if we are not careful.

\section{Amazon Mechanical Turk Setup}
\begin{figure}
    \centering
    \includegraphics[width=2.0\linewidth]{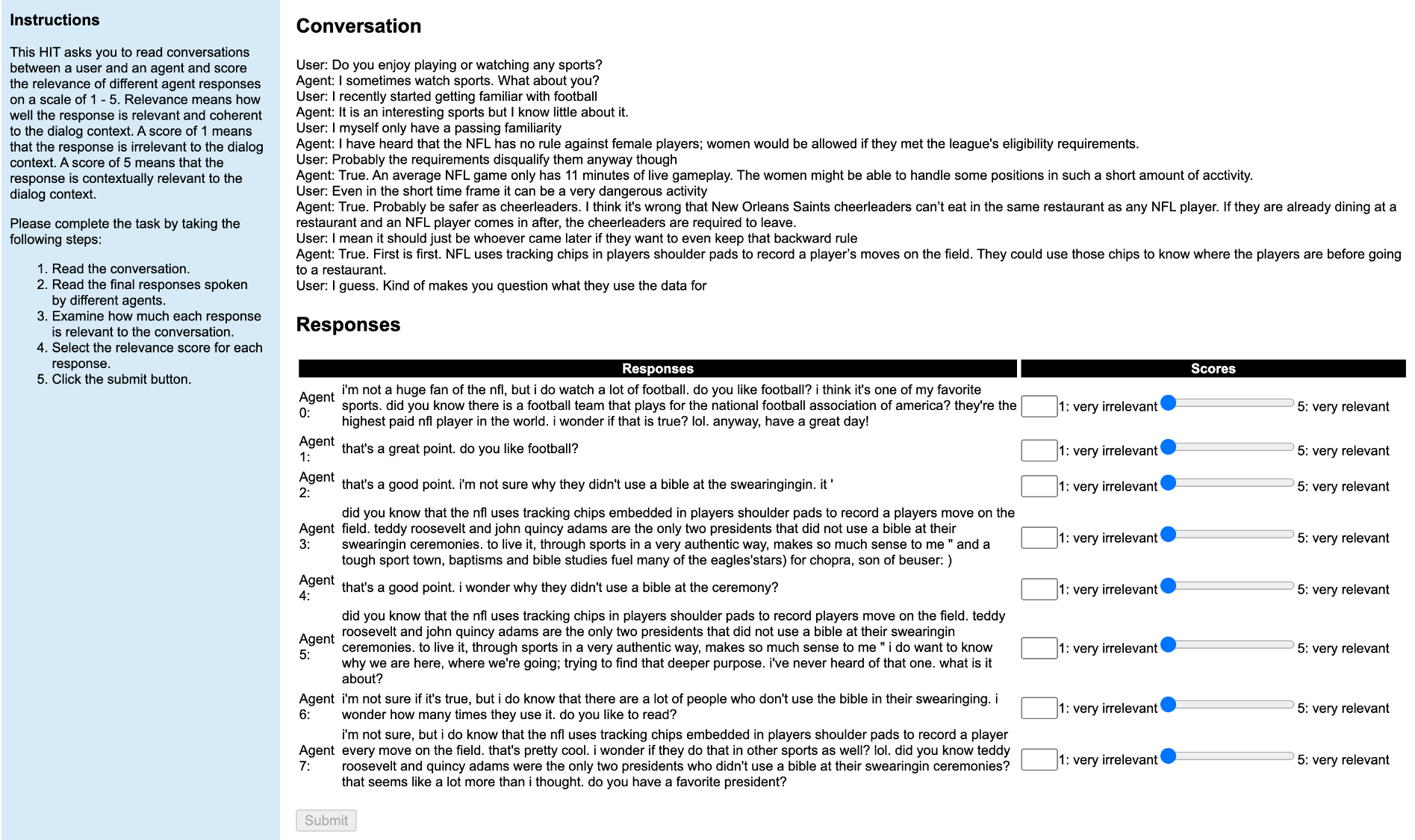}
    \caption{Our human evaluation setup.}
    \label{fig:amt_setup}
\end{figure}

In Figure \ref{fig:amt_setup} we show a screenshot of our Amazon Mechanical Turk setup for human evaluation.

\end{document}